# A Deep Learning Bidirectional Temporal Tracking Algorithm for Automated Blood Cell Counting from Non-invasive Capillaroscopy Videos

Luojie Huang, Gregory N. McKay, and Nicholas J. Durr

*Abstract*—Oblique back-illumination capillaroscopy has recently been introduced as a method for high-quality, non-invasive blood cell imaging in human capillaries. To make this technique practical for clinical blood cell counting, solutions for automatic processing of acquired videos are needed. Here, we take the first step towards this goal, by introducing a deep learning multi-cell tracking model, named CycleTrack, which achieves accurate blood cell counting from capillaroscopic videos. CycleTrack combines two simple online tracking models, SORT and CenterTrack, and is tailored to features of capillary blood cell flow. Blood cells are tracked by displacement vectors in two opposing temporal directions (forward- and backward-tracking) between consecutive frames. This approach yields accurate tracking despite rapidly moving and deforming blood cells. The proposed model outperforms other baseline trackers, achieving $65.57\%$ Multiple Object Tracking Accuracy and $73.95\%$ ID F1 score on test videos. Compared to manual blood cell counting, CycleTrack achieves $96.58 \pm 2.43\%$ cell counting accuracy among 8 test videos with 1000 frames each compared to $93.45\%$ and $77.02\%$ accuracy for independent CenterTrack and SORT almost without additional time expense. It takes 800s to track and count approximately 8000 blood cells from 9,600 frames captured in a typical one-minute video. Moreover, the blood cell velocity measured by CycleTrack demonstrates a consistent, pulsatile pattern within the physiological range of heart rate. Lastly, we discuss future improvements for the CycleTrack framework, which would enable clinical translation of the oblique back-illumination microscope towards a real-time and non-invasive point-of-care blood cell counting and analyzing technology.

L. Huang is with the Department of Biomedical Engineering, Johns Hopkins University, Baltimore, MD 21218 USA. (e-mail: lhuang48@jhu.edu)

G. N. McKay is with the Department of Biomedical Engineering, Johns Hopkins University, Baltimore, MD 21218 USA. (e-mail: gmckay1@jhu.edu)

Corresponding author: N. J. Durr, is with the Department of Biomedical Engineering, Johns Hopkins University, Baltimore, MD 21218 USA. (e-mail: ndurr@jhu.edu)

## INTRODUCTION

COMPLETE blood count (CBC) is the most common clinical test with approximately 34.5 million tests performed annually in the United States [1]. The CBC measures the concentration of cells in a patient's blood, including white blood cells (WBCs), red blood cells (RBCs), and platelets (PLTs). The CBC is useful in managing a wide range of conditions and diseases, such as cancer, anemia, infection, and inflammation, as well as for prognostic monitoring of chemotherapy and radiation therapy [2].

The current clinical CBC is administered invasively through venipuncture in adults and heelstick in newborns. The drawn blood is processed via a hematologic analyzer, which typically analyzes the scattering signal of each cell in a flow cytometer [3]. While the CBC is a ubiquitous and largely effective clinical test, some vulnerable patients, such as immunocompromised cancer patients who are susceptible to infection and neonates who are susceptible to phlebotomy-induced anemia, would benefit from a non-invasive technique [4], [5]. An accurate, non-invasive CBC would significantly enhance treatment efficiency and reduce unnecessary risks for such patients.

In recent years, there are emerging advances towards minimal-invasive and non-invasive point-of-care blood cell analysis technology. Hemocue (Ängelholm, Sweden) is a portable hemoglobinometers that can provide fast and reliable Hemoglobin concentration results with only $10\mu L$ blood from finger prick [6]. Leuko Labs is translating a technology called the PointCheck, with the goal of non-invasive screening for severe neutropenia in chemotherapy patients. This approach relies on an inverse correlation between degree of neutropenia and the frequency of optical absorption gaps observed in nailfold capillary videos. Based on their clinical study, event counting of absorption gaps within one-minute videos allows valid discrimination between severe neutropenia and baseline [7]. However, it is challenging to obtain images with resolvable blood cells from nailfold capillaries due to their depth beneath highly scattering skin.

Recently, McKay *et al.* introduced oblique back-illumination capillaroscopy (OBC), an affordable technique that acquires high-quality phase and absorption contrast images of blood cells flowing in tongue capillaries *in vivo* [8]. OBC works regardless of skin tone and can be implemented with a simple camera, microscope objective, and LED. To enable OBC to complete a non-invasive blood counting and analysis, there is a need for efficient, automated, and accurate OBC video analysis algorithms. Based on the unique features of blood cell flow in OBC videos, we present a deep multi-object tracking framework named CycleTrack that accurately tracks and counts blood cells flowing through human capillaries *in vivo*.

## I. RELATED WORK

### A. Oblique back-illumination capillaroscopy

OBC is a non-invasive, label-free method for high-speed *in-vivo* blood cell imaging. Blood cells in superficial capillaries can be clearly visualized using phase contrast generated by an illumination-detection offset, and hemoglobin absorption contrast generated by green LED illumination. A simple and compact OBC system has been developed that achieves 200 Hz imaging of a $250 \times 250 \ \mu m^2$ field of view, demonstrating resolvable RBCs, WBCs, and PLTs in human ventral tongue capillaries, which are normally $70 - 80 \ \mu m$ below the surface. The system simplicity, robustness, imaging speed, and quality makes it a promising technology for non-invasive blood counting and analysis. OBC imaging can capture the images of thousands of blood cells in minutes. Therefore, there is a need for video analysis algorithms to efficiently process OBC data and achieve an accurate cell count.

### B. Multiple Object Tracking (MOT)

For video object counting task, counting-by-detection is the most intuitive method. in addition to counting, this method can also extract individual objects for extended studies. To count cells from video measurements, the MOT model, a typical counting-by-detection method, can be used to identify and track individual cells across frames. State-of-the-art MOT algorithms utilize Deep Learning based object detectors such as Fast-RCNN [9], MaskRCNN [10], and CenterNet [11]. The approach employed in these models follows the tracking-by-detection pipeline, which first identifies objects in each frame with segmented masks or bounding boxes, and then associates detected objects across frames based on appearance features, location or topological information. Among these frameworks, appearance-based tracking is a particularly intense area of study thanks to powerful convolutional feature extractors such as a conventional convolutional neural network [12] or ResNet [13].

However, unlike objects in most benchmark MOT datasets, cells of a given class in a capillaroscopic video have similar appearance—they are approximately the same size, shape, color, and granularity. Moreover, blood cells are deformable, and their shapes and appearances tend to change frame-to-frame due to unpredictable factors as they flow through crowded capillaries, due to rotation, collision, and movement in and out of the focal plane. Therefore, the task of capillaroscopic cell tracking is unique and challenging, as it is difficult to distinguish and assign a specific trajectory to individual blood cells using off-the-shelf appearance-based MOT models. To reliably track blood cells within these constraints, we introduce a novel architecture, called CycleTrack, which combines advantages of two simple yet effective real-time trackers with frame-to-frame inputs, named online tracking (CenterTrack [14] and SORT [15]), by cyclically applying them in forward- and backward-directions to video data.

### C. Tracking Objects as Points: CenterTrack

CenterTrack is an online spatiotemporal algorithm for tracking multiple detected objects across sequential frames. It uses a one-stage object detector, CenterNet, which detects each object as a triplet, estimating its center and the size of its bounding box [14]. CenterTrack takes two consecutive frames with previously tracked objects as input, and outputs the same triplet as CenterNet with an additional displacement vector from the current frame to the previous frame. These displacement vectors can be used to translate new detection centers to previous states to match existing objects by a simple greedy matching method. Errors of CenterTrack mainly occur from detection errors and biases of displacement vectors [14].

### D. Simple Online and Realtime Tracking (SORT)

SORT is another online spatiotemporal tracker that employs an assumption that each individual object's movement can be modeled as a kinetic system with a predictable pattern. Based on objects' bounding boxes extracted by standard detectors, SORT efficiently associates objects in real-time by propagating an object's state into future frames using Kalman filters [16]. To associate current detections with existing objects, SORT uses either the Hungarian algorithm [17] to optimize overall Intersection over Union (IoU) of all bound boxes, or the greedy matching method to find the best match with closest center distances. As SORT is based on simple movement assumption, it is sensitive to velocity changes, which are quite common in blood flow videos because of cardiac cycle. Additionally, due to the simplicity of the tracking algorithm, SORT enables tracking at high speeds, which will be crucial for real-time capillaroscopic analysis.

SORT tracks in a forward manner, assuming that objects move in a predictable and continuous pattern over time and space, while CenterTrack works in a retrospective way globally matching a constellation of object centers in the current frame to the previous. Therefore, the conceptual difference and opposing tracking directions make the cooperation of two trackers likely to contribute to a more robust tracking scheme. In OBC videos, blood cells move in fixed directions along capillaries, which approximately meets the SORT assumption of predictable velocity. Capillaroscopic blood cell tracking is also an appropriate use case for CenterTrack, because in crowded capillaries, the relative positions between nearby cells tend to remain consistent during capillary flow. Following this intuition, CycleTrack combines SORT and CenterTrack into a robust tracking algorithm that tracks objects in both temporal directions.

## II. METHODS

CycleTrack combines SORT and CenterTrack into a robust tracking algorithm that tracks objects in both temporal directions. The CycleTrack framework is shown in Fig. 1. CycleTrack combines CenterTrack and SORT to achieve backward and forward tracking between two consecutive frames. To provide a more detailed CycleTrack workflow. Algorithm 1 outlines the tracking process step by step. In this section, the description of CycleTrack is organized around

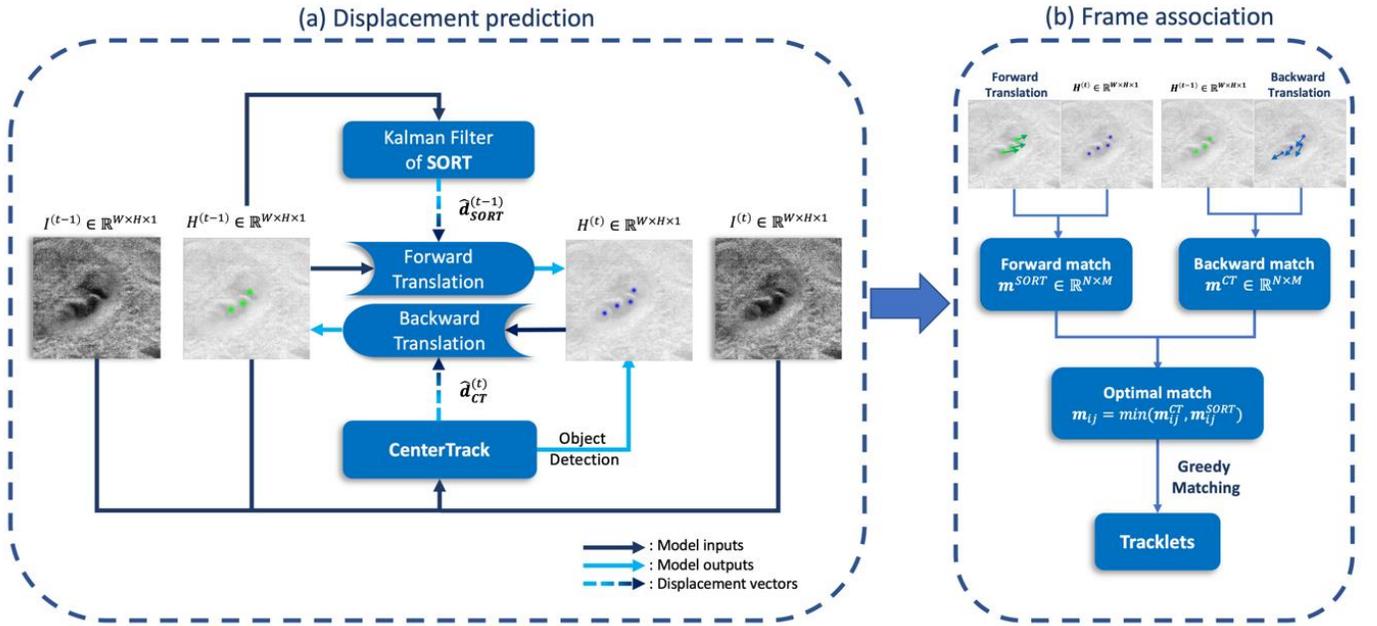

**Fig. 1. Workflow of CycleTrack Framework.** CycleTrack performs tracking in two steps: displacement prediction and frame association. (a) Displacement prediction: CycleTrack predicts displacement between consecutive frames in two temporal directions: forward translation and backward translation. (i) Forward translation: SORT takes detected object centers from the previous frame $H^{(t-1)}$ as inputs to update the Kalman filter and predicts displacement from the previous frame to the current frame. (ii) Backward translation: CenterTrack takes the previous frame image $I^{(t-1)}$, detected object centers in the previous frame $H^{(t-1)}$ and the current frame image $I^{(t)}$ as inputs, and predicts object centers in the current frame $H^{(t)}$ (which will be used at $t+1$) and displacement from the current frame to the previous frame. (b) Frame association: The optimal matching plan is decided by the closest distances between centers in two frames using the greedy matching algorithm.

---

**Algorithm 1:** Workflow of CycleTrack

**Input:** Input OBC video frames $I = \{I^{(j)} \in \mathbb{R}^{W \times H \times 1}\}_{j=1}^{T}$

**Output:** Tracked object list $Tr$

1    // **Initialization:** Tracked object list $Tr$ as CenterTrack detections in first frame and base vector $\hat{d}_{base}$ as empty list.
2    $Tr^{(1)} \leftarrow \text{CT}(I^{(1)})$ // CT($\cdot$) means detection outputs from CenterTrack
3    $\hat{d}_{base} \leftarrow \emptyset$ // empty list
4    **for** $t \leftarrow 2\ to\ T$ **do**
5      // object detection $D^{(t)} = \{(p, s, \omega, id)_j\}_{j=1}^{M}$ and backward displacement
6      $D^{(t)}, \hat{d}_{CT}^{(t)} \leftarrow \text{CT}(I^{(t-1)}, I^{(t)}, Tr^{(t-1)})$
7      **if** $\hat{d}_{base}^{(t-1)} \neq \emptyset$ **do**
8        $\hat{d}_{CT}^{(t)} \leftarrow w\hat{d}_{CT}^{(t)} + |1 - w|\hat{d}_{base}^{(t-1)}$ // base vector correction
9      **end**
10      // Forward tracking: SORT
11      $\hat{d}_{SORT}^{(t-1)} \leftarrow \text{SORT}(Tr^{(t-1)})$ // forward displacement
12      // optimal matching plan
13      $m_{ij}^{CT} \leftarrow \left| p_i^{(t-1)} - (p_j^{(t)} + \hat{d}_{CT\ j}^{(t)}) \right|$ // CenterTrack cost matrix: $m^{CT} \in \mathbb{R}^{N \times M}$
14      $m_{ij}^{SORT} \leftarrow \left| p_j^{(t)} - (p_i^{(t-1)} + \hat{d}_{SORT\ i}^{(t-1)}) \right|$ // SORT cost matrix: $m^{SORT} \in \mathbb{R}^{N \times M}$
15      $m_{ij} \leftarrow min(m_{ij}^{CT}, m_{ij}^{SORT})$ // minimal cost matrix
16      $Tr^{(t)} \leftarrow \text{GM}(Tr^{(t-1)}, m)$ // greedy matching
17      // update base vector
18      $\hat{d}_{base}^{(t)} \leftarrow mean(\hat{d}^{(t)})$ // average of displacements from optimal plan
19    **end**
20    **Return:** $Tr$

---

its three key components of CenterTrack, SORT, and association of new cell detections with the existing tracks of previous cells, which are termed as tracklets.

### A. Cell detection and backward tracking via CenterTrack

CenterTrack is a single deep network that can solve object detection and tracking jointly and is trained end-to-end CenterTrack uses a CenterNet [11] detector which takes a single image $I \in \mathbb{R}^{W \times H \times 1}$ as the input and outputs object detections. Each output detection $y = (p, s, \omega, id)$ is represented by its center location ($p \in \mathbb{R}^2$), height and width of the bounding box ($s \in \mathbb{R}$), a confidence score ($\omega \in [0,1]$) and a detection id ($id \in \mathbb{Z}, id > 0$).

With the Deep Layer Aggregation model (DLA) [18] as the network backbone, the architecture of CenterTrack is nearly identical to CenterNet. Compared with the original CenterNet, CenterTrack simply expands the input and output channels to achieve multiple tasks. CenterTrack takes two consecutive frames: the current frame $I^{(t)} \in \mathbb{R}^{W \times H \times 1}$ and the prior frame $I^{(t-1)} \in \mathbb{R}^{W \times H \times 1}$, and the prior tracked objects $Tr^{(t-1)} = \{y_{tracked\,0}^{(t-1)}, y_{tracked\,1}^{(t-1)}, ...\}$ as inputs. To represent detections in a form that could be easily input into CenterTrack, all the detections in $Tr^{(t-1)}$ are rendered in a ground-truth heatmap $H^{(t-1)} \in \mathbb{R}^{W \times H \times 1}$, within which each detection is represented by a Gaussian-shaped point. Based on the rendering function, the intensity at every position $q \in \mathbb{R}^2$ in the heatmap is defined as:

$$H^{(t-1)}(q) = \max_i exp\left(-\frac{(p_i^{(t-1)}-q)^2}{2\sigma_i^2}\right), \quad (1)$$

Here $p_i^{(t-1)}$ is the location of the $y_{tracked\,i}^{(t-1)}$, and $\sigma_i$ is defined as 1/3 of the mean of the width and height of its bounding box [19].

Besides detections in the current frame $Y^{(t)} = \{y_0^{(t)}, y_1^{(t)}, ...\}$, CenterTrack also predicts a 2D displacement map $\widehat{D}^{(t)} \in \mathbb{R}^{W \times H \times 2}$ as 2 additional output channels. The displacement vectors $\hat{d}^{(t)}$ representing the predicted cell's lateral movement as location change from the current frame to the previous frame, are learned using L1 regression:

$$L_{off} = \frac{1}{N}\sum_{i=1}^{N}\left|\hat{d}_i^{(t)} - (P_i^{(t-1)} - P_i^{(t)})\right|, \quad (2)$$

where $P_i^{(t-1)}$ and $P_i^{(t)}$ are ground-truth detection locations. To restrict displacement estimations to adhere to the assumption that all blood cells in the same frame should move in similar directions, we introduce a base vector $\hat{d}_{base}^{(t-1)}$, which is the average displacement vector of all cells from frame $(t-1)$. This base vector is used to refine displacement vector predictions from CenterTrack, $\hat{d}_{CT\,i}^{(t)}$:

$$\hat{d}_{CT\,i}^{(t)} = w_i \hat{d}_i^{(t)} + |1 - w|\hat{d}_{base}^{(t-1)}, \quad (3)$$

where $w_i = \frac{\hat{d}_{base}^{(t-1)} \cdot \hat{d}_i^{(t)}}{\left|\hat{d}_{base}^{(t-1)}\right|\left|\hat{d}_i^{(t)}\right|}$. This equation provides a weighted, corrective action on the conventional displacement vector prediction from CenterTrack. The more $\hat{d}^{(t)}$ deviates from $\hat{d}_{base}^{(t-1)}$, the more the refined vector ($\hat{d}_{CT}^{(t)}$) would rely on base vector ($\hat{d}_{base}^{(t-1)}$).

### B. Forward tracking via SORT

SORT is an unsupervised tracking model, approximating each object's displacement from the previous frame to the current frame with a linear constant velocity model. This is accomplished with a Kalman filter, which is commonly used for state transition prediction in linear dynamic systems [20]. A Kalman filter is created for each tracklet, and to update the Kalman filter, the input state of each tracked object is modeled as $d_{SORT} = (p_{SORT}, s_{SORT})$, where $p_{SORT} \in \mathbb{R}^2$ is the object's center and $s_{SORT} \in \mathbb{R}$ is the bounding box size. Finally, the state transition $\hat{d}_{SORT}^{(t-1)}$ from the previous frame to the current frame is predicted for tracking by:

$$\hat{d}_{SORT}^{(t-1)} = p_{SORT}^{(t)} - p_{SORT}^{(t-1)}, \quad (4)$$

### C. Association between consecutive frames

Our implementation of CenterTrack outputs object displacement vectors pointing from the current frame to the previous frame. Using these displacements, we simply translate new detection centers backwards to the previous frame. The matching cost matrix $m^{CT} \in \mathbb{R}^{N \times M}$ is then computed as the Euclidean distances between the centers of N tracked objects and M translated detections. For the $i_{th}$ tracked object and the $j_{th}$ translated detection:

$$m_{ij}^{CT} = \left|p_i^{(t-1)} - (p_j^{(t)} + \hat{d}_{CT\,j}^{(t)})\right|, \quad (5)$$

From SORT, we get another matching cost matrix $m^{SORT} \in \mathbb{R}^{N \times M}$ between $N$ predicted locations of objects from the previous frame and $M$ new detections in the current frame:

$$m_{ij}^{SORT} = \left|p_j^{(t)} - (p_i^{(t-1)} + \hat{d}_{SORT\,i}^{(t-1)})\right|, \quad (6)$$

To combine the matching estimates from CenterTrack and SORT between consecutive frames, we first select the smaller distance for each element in the matching cost matrix $m_{ij} = min(m_{ij}^{CT}, m_{ij}^{SORT})$. Then, a greedy matching algorithm is applied to match detections to the tracked objects with the closest mutual distances. Moreover, as an additional restriction, if all the distances of a detection are further away than a reasonable threshold, it will be regarded as unmatched and a new tracklet will be created for this newly detected cell. The threshold is set adaptively to be the average distance between adjacent cells in the same frame.

## III. EXPERIMENTS

We trained and evaluated the proposed model on videos of human ventral tongue capillaries acquired by the OBC system with approval by the Johns Hopkins University Institutional Review Board (IRB00204985). Videos were acquired by the oblique back-illumination capillaroscopy. The OBC system is fully described in [8] and uses a Green LED ($527\ nm$) as the light source with an illumination-detection offset of around $200\ \mu m$ (in object space) (Fig. 2). Using a 40x 1.15 NA water immersion microscope objective, videos with a frame size of $1280 \times 812$ pixels were acquired at $160\ Hz$ and 0.5ms exposure time, with a $416 \times 264\ \mu m^2$ field of view.

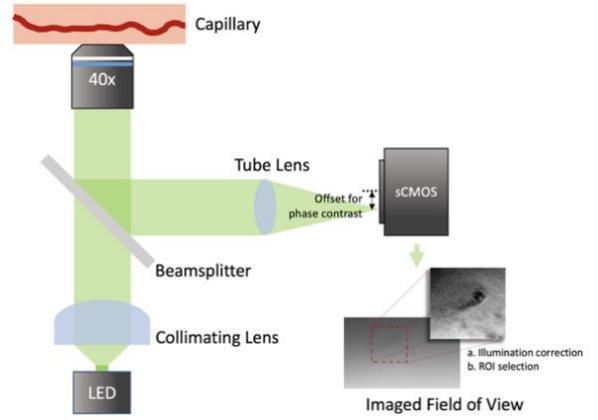

**Fig. 2. Setup of the OBC system.** A green LED is placed at the focal plane of a collimating lens and imaged critically into a capillary bed by a 40x objective lens. Back-scattered light is collected by the objective, reflected off a beamsplitter and imaged onto an offset sCMOS detector by a tube lens. This lateral offset reduces direct backscattered light and enhances phase contrast from obliquely remitted light.

## A. Data Preparation

We acquired videos from 4 different ventral tongue capillaries with different shapes and cell densities. During model hyperparameter tunning, we applied 4-fold cross validation by selecting a 100-frame fully annotated video from each capillary. And the video data were split based on capillaries to prevent capillary feature leakage. And with the optimal hyperparameters, the final model is trained on videos from 3 capillaries while videos from another capillary were left for test. The training dataset contains 335 fully annotated frames from 3 different sequences with a total of 2280 masks for 151 cells. Each annotation was created manually with a labeled mask and a tracking ID. All tracking IDs are consistent across frames for the same cell in a sequence. The testing dataset contained a sequence with 300 annotated frames from a new capillary, and traditional MOT evaluation metrics were computed. For further validation of cell count estimation accuracy, we applied CycleTrack to eight additional videos with 1000 frames each. These videos had manually determined cell counts as ground truth but no mask annotations.

## B. Video Stabilization & Region Selection

This section introduces the video pre-processing steps in our experiments. During *in-vivo* imaging, video instability caused by natural micron-level human motion introduces additional difficulty to the object tracking algorithm. Thus, as the first step of preprocessing, we apply automatic video stabilization using background point feature matching [21]. This method tracks feature points from the background texture in two consecutive frames. The tracked points enable us to estimate background motion, including rotation, vertical and horizontal translations, and compensate for them. A flow estimation is applied throughout each stabilized video by simply accumulating pixel value changes between consecutive frames, where the capillary lumen tends to have higher intensity due to flowing blood cells. To increase tracking efficiency and to reduce detection errors from the background, a $512 \times 512$ pixel region of interest (ROI) was selected by centering the capillary based on the flow estimation results. Finally, we mitigate differences from variation in illumination setups and experiments by applying contrast limited adaptive histogram equalization (CLAHE) [22] and standard normalization of pixel intensity to each image.

## C. Implementation Details

CycleTrack builds upon the CenterNet based CenterTrack and SORT. For CenterTrack, we used a modified DLA model [18] as a backbone. The training inputs were made up of frame pairs. To make the model generalize to different capillaries, training data augmentation, including rotation, vertical and horizontal flips, and temporal flips on frame pairs were applied to simulate blood cell flow in a wide variety of directions. Moreover, frame pairs were generated with various frame differences in the range of $[-3,3]$ to simulate up to 3x faster flow velocities. During training, we used the focal loss in the original CenterNet work [11] for object detection and offset loss $L_{off}$ for displacement vector regression, optimized with Adam with a learning rate of $10^{-4}$ and batch size of 16. For data augmentation, we set the probability of vertical/horizontal flips to 0.5 and rotations to be uniformly varying within 15 degrees of the original orientation. For all experiments, we trained the networks from scratch for 300 epochs. The learning rate was reduced by half every 90 epochs. We tested CycleTrack runtime on an Intel Xeon E5-2620 v4 CPU with a Titan V GPU. We only track detections with a confidence of $\omega \geq 0.6$. During CenterTrack training, test-time detection errors simulation [14] is an effective strategy to better tolerate the imperfect object detector at inference time, thereby improving the final tracking accuracy. To simulate errors from detections, we set random false positive and false negative ratios as $\lambda_{fp} = 0.1$ and $\lambda_{fn} = 0.4$, respectively. All the thresholds were tuned by the aforementioned cross validation experiments.

## D. MOT Evaluation Metrics

We compute the official evaluation metrics from MOT benchmarks [23,24] on the test dataset, which comprehensively evaluate model performance in both object detection and tracking. These metrics mainly evaluate MOT in four aspects, including: detection, ID match, trajectory, and overall accuracy. In evaluation measures labeled with (↑), higher scores denote better performance, while for evaluation measures labeled with (↓), lower scores denote better performance. We compared our the CycleTrack with several benchmark online MOT models including Tracktor++ v2 [25], MaskTrack [26], SORT, and CenterTrack. Tracktor++ v2 is an effective bounding box regressor that tracks by predicting tracked objects in future frames and uses an extensive re-identification (Re-ID) model to achieve stable tracking over a long time range. In our experiments, Tracktor++ v2 is trained on our dataset following default configurations in [25] using Faster R-CNN with feature pyramid network as the object detector. MaskTrack is another multi-object tracker built upon the powerful Mask R-CNN object detector. Following the same tracking-by-detection concept, MaskTrack finds object associations across frames by calculating similarities of extracted appearance features. For the SORT model, we used object detections from CenterNet. for a comparative model focused on the tracking task. For a fair comparison, we applied the same aforementioned pre-processing process to videos before different trackers.

## IV. RESULTS AND DISCUSSIONS

### A. MOT Evaluation

Fig. 3 shows example inputs and outputs of CycleTrack over 5 consecutive frames. In these qualitative assessments, we observe CycleTrack is successfully detecting and tracking blood cells frame-to-frame. A quantitative analysis of performance is presented in Table 1, where we list the MOT metrics of CycleTrack and other comparative models tested on an unseen 300-frame video with full annotations. In terms of object detection, thanks to the simple morphology

Table 1: **MOT Metrics on Comparative Models. A comparison between baseline models and Proposed CycleTrack. All results are on the test OBC videos.**

| Model | Detection | | ID match | | | Trajectory | | | | Overall | Speed |
|---|---|---|---|---|---|---|---|---|---|---|---|
| | Prcn↑ | Rcll↑ | IDP↑ | IDR↑ | IDF1↑ | MT↑ | ML↓ | IDSw↓ | Frag↓ | MOTA↑ | (Hz) |
| Tracktor++ v2 | 88.29 | 83.88 | 63.6 | 57.33 | 61.10 | 31.14 | 24.83 | **26.67** | **15.69** | 46.36 | 1.5 |
| MaskTrack | 90.28 | **85.75** | 67.17 | 64.52 | 65.92 | 41.23 | 22.92 | 38.10 | 24.46 | 53.22 | 2.6 |
| SORT | **92.16** | 82.42 | 68.5 | 64.0 | 66.13 | 51.7 | 18.49 | 69.06 | 43.39 | 55.82 | **258** |
| CenterTrack | **92.16** | 82.42 | 72.12 | 71.25 | 71.64 | 65.83 | 15.54 | 61.15 | 30.61 | 61.61 | 12 |
| CycleTrack (Ours) | **92.16** | 82.42 | **75.21** | **72.74** | **73.95** | **71.43** | **13.27** | 36.34 | 22.93 | **65.57** | 12 |

* Prcn: precision; Rcll: recall; IDP/IDR: ID precision / recall; IDF1: ID F1 score; MT / ML: percentage of most tracked/lost trajectory; IDSw: percentage of an ID switches to a different previously tracked object; Frag: percentage of fragmentations where a track is interrupted by miss detection; MOTA: multi-object tracking accuracy.

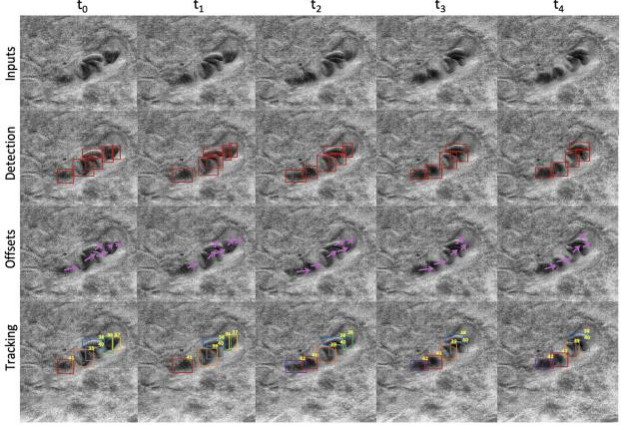

**Fig. 3. Example of CycleTrack Outputs.** The first row shows the original inputs at consecutive frames of times ($t_0 - t_4$). The second row shows the bounding boxes predicted by the object detector (CenterNet) from CycleTrack. The third row shows forward displacement vectors of the optimal matching plan by CycleTrack. The last row shows the final tracking results. Tracked cells in different frames would be assigned to an identical tracking ID.

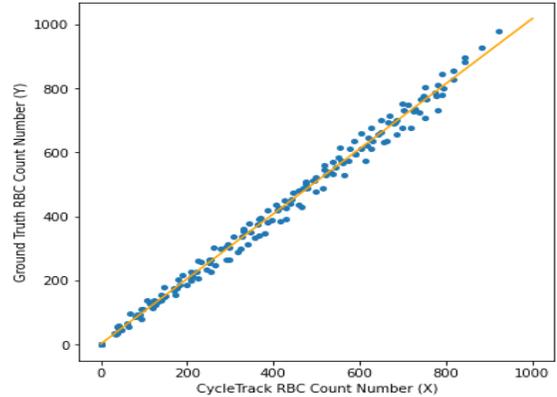

**Fig. 4. Correlation between the ground truth blood cell count and results from CycleTrack.** $Y = 1.0186X + 1.0081$ (Y = ground truth and X = cycleTrack count). The correlation coefficient ($\gamma$) within 1000 frames count is 0.9960. These results indicate a very strong correlation between CycleTrack count and the ground truth.

and good phase and absorption contrast, the cells in the OBC videos can be easily discriminated from the background by all the tested object detectors. Therefore, all 5 models achieved competitive accuracy in the detection task with precisions above 88%. Since all comparative models achieve sufficient detection accuracy, we can fairly compare their tracking performance in the tracking-by-detection pipeline. From the detection metrics, we observe that recall is lower than precision across all models, which means that errors were more commonly from false negatives. Unsurprisingly, we observe that most detection errors result from the failure to detect overlapping cells in crowded capillaries. Segmenting overlapping cells was also the most difficult part of manually labeling the ground truth data. In the tracking metrics, we find that CycleTrack outperforms all other trackers on OBC videos, setting a new state-of-the-art performance in terms of multiple object tracking accuracy (MOTA), which provides the most intuitive general evaluation of object detection and tracking by taking both detection errors and matching errors of tracking into account, and ID F1 score (IDFl). We also notice that, compared with the online trackers without Re-ID models, like SORT and CenterTrack, the major improvement of our framework is in the reduction of ID switches, which we measured to be reduced by approximately half. ID switches have widely shown to be a good metric that reflects the capability of measuring long, consistent tracks [27, 28]. High frequency of ID switches is typically an inherent problem for local online trackers without Re-ID models, like SORT and CenterTrack. Since such trackers only focus on associations within two consecutive frames, a missed detection or biased displacement vector would cause an irreparable break of tracklets that leads to overcounting. Unlike other Re-ID models that refine long tracklets in a wider time range, CycleTrack remains focused on just two consecutive frames. We believe the back-and-forth tracking paths in CycleTrack compensate for each other's errors to find a better matching plan in consecutive frames. The reduction of ID switches demonstrates the effectiveness of our combined framework to stabilize tracking by optimizing cell associations in adjacent frames on our OBC videos.

*B. Counting Results*

We also evaluate the cell counting accuracy of CycleTrack on 8 different long videos, each containing 1000 frames with manually counted ground truth (but no ground truth segmentation labels). The agreement between CycleTrack count and ground truth is shown Fig. 4. The correlation coefficient ($\gamma$) calculated from these experiments is 0.9960 which indicated a very strong positive relation between CycleTrack count and the ground truth. Fig. 5 shows the percentage of counting error as more frames are analyzed on these 8 videos. As frame number increases, we observe a descending trend in both average and variance of errors among different videos. Throughout the 1000 frame videos, the absolute counting errors decreased quickly at first 300 frames and then became more stable as the base vector from CenterTrack gradually found the right direction of capillary flow. When the frame number exceeds 600, the model typically has counted more than 400 cells, and the average count error stabilizes below 5%. This indicates that, when analyzing single capillaries in clinical scenarios, videos should at least have 600 frames (3.75 seconds) to get a reliable counting result with CycleTrack. By 1000 frames (6.25 seconds of video), the CycleTrack method reaches a state-of-the-art average counting accuracy of 96.58%, compared to 93.45% and 77.02% accuracy of original CenterTrack and SORT respectively. This demonstrates that, with the same object detector, stabler long-time tracking resulting from reduction in ID switches and broken trajectories (fragments) contributes to a great improvement to the final counting accuracy. Our frame-by-frame analysis of the source of these errors revealed that they are mainly caused by missed detections, which can lead to either under-counting for cells that are totally missed or over-counting from tracklet breaks. Because these failure modes offset and happen at similar frequencies, they neutralize each other, thus our counting accuracy is better than the MOTA metrics suggest. The absolute count errors are expected to be sensitive to sudden changes of blood flow velocity and cell density. However, CycleTrack can overcome these challenges because the tracker is constantly updated by the latest tracking states. Though short fluctuations in flow velocity do affect performance, the longer periods of slow-changing flow rates allow for accurate counting as the tracking model adapts to these intermittent but stable states.

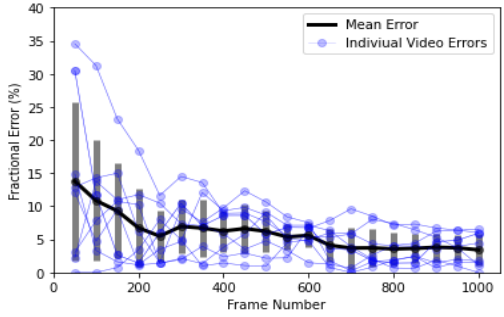

**Fig. 5. Fractional Counting Errors Across Frames.** CycleTrack is tested on 8 different 1000-frame OBC videos. Each blue curve represents the absolute percentage counting error for an increasing number of frames from a single video. Counting errors are reported every 50 frames. The black curve are the mean, and grey bars are the standard deviations of count errors computed from all 8 videos.

### C. Runtime

The runtime mainly depends on the input image resolution and the number of objects detected and tracked. With a 16-

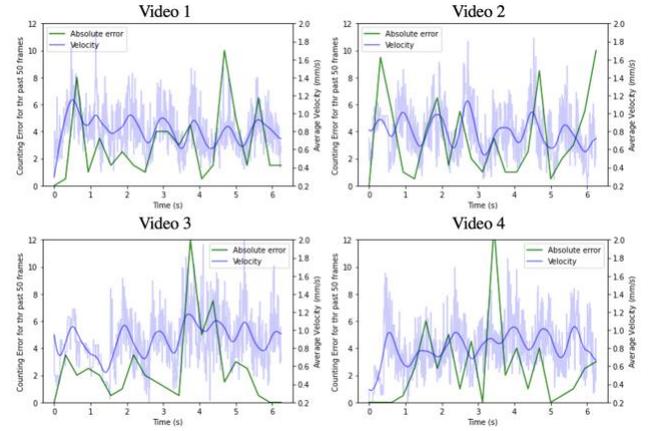

**Fig. 6. Velocity Estimation and Absolute Counting Errors Over 4 Different Test Videos.** The faded blue lines show the average velocity of all objects detected in each frame from CycleTrack. The solid blue lines show a lowpass filtered average velocity highlighting the oscillatory rhythm of blood flow at the capillary level. The green lines show the absolute cell counting errors for the past 50 frames compared to manually counted ground truth. Most peaks of errors align with the velocity peaks, indicating the count error rates are correlated to blood velocity.

bit image of $512 \times 512$ pixels in size, CycleTrack ran at 12 frames per second. The runtime of comparative models is shown in the last column in Table 1. Thanks to the efficiency of CenterNet, CycleTrack has a great advantage over other trackers in terms of speed while maintaining a good detection accuracy. Moreover, as SORT is a fast, unsupervised online tracking model capable of real-time tracking at up to 258 Hz, incorporating this model in CycleTrack adds no significant computational costs over the original CenterTrack. For each frame, CycleTrack takes ~$80ms$ for detection and ~$5ms$ for tracking, making it 14x slower than the frame acquisition rate (160 Hz). It takes $800s$ to track and count approximately 8000 blood cells from 9,600 frames captured in a typical one-minute video, which is significantly faster than Tractor++ and MaskTrack. However, compared to the imaging rate, the current CycleTrack still need further improvements, especially more efficient object detectors, in order to achieve actual real-time tracking.

### D. Extracting heart rate and assessing performance correlation for velocity

Since the errors are from the previous 50 frames, we believe that the velocity increase will cause an increase in counting errors. Fig. 6 shows four examples of the estimated average velocity across frames from predicted displacement vectors with the absolute errors of the past 50 frames. From these data, we observe a clear sinusoidal signal at a frequency of approximately 1 Hz. This falls within the expected normal physiological heartbeat of around 60 beats per minute. Therefore, we believe it may be possible to assess other significant physiological information beyond blood counts with this technology, such as blood flow metrics and vital signs. Moreover, these data show that the absolute error peaks tend to align with the peak blood flow velocity, which indicates a correlation between counting errors and flow velocity.

*E. Discussion*

A verification study by Vis *et al.* [29] reported that the state-of-the-art analytical blood cell count accuracy of routine hematology analyzes is above 96.8%. And compared with hematology analyzer, the average percent coefficient of variation (CV) of fingerprick blood cell counts from point-of-care instruments was higher by at least 3 times [30], which suggests that current point-of-care blood counting techniques achieve clinical impact with the sacrifice of report robustness and accuracy. Compared with manual counting results, the proposed CycleTrack achieved state-of-the-art blood cell tracking and an average accuracy of 96.58% on 1000-frame OBC videos using manual counting as the ground truth. However, the OBC system is still in an exploratory research stage and there is no existing data assessing the accuracy of a manual blood count from video frames with blood counts per volume determined by a gold standard venous blood draw and flow cytometry. The time required for acquiring and processing the 600 frames required for 95% agreement was 3.75 seconds of video recording at 160 Hz, followed by 51 seconds to complete CycleTrack analysis. As conventional and point-of-care blood analysis requires a blood draw or finger prick, it is likely that this time for measurement can be acceptable for several clinical use cases. There are many opportunities for future work. Firstly, different cell types can be qualitatively distinguished under OBC. For example, WBCs are brighter and bigger than RBCs, while platelets are smaller than both RBCs and WBCs. With an expanded training dataset, cell classification could be explored using the object detector from the current model. A clinical study on the superficial capillaroscopy video has shown a promising result on severe neutropenia detection by counting white blood cells in 2-3 one-minute imaging sessions for each patient [31]. Therefore, multiple cell-class tracking would enable evaluating more of the cellular components of blood, moving closer to a non-invasive CBC, and with more clinical study, would allow the non-invasive screening of more physiologic states such as neutropenia, inflammation, and thrombocytopenia. The bottleneck for this task may be the acquisition of large datasets with reliable ground truth blood cell type annotations, especially for the rarer blood cell types. Secondly, to calculate absolute cell concentration, it is necessary to compute flow rates and volumes of each capillary segment. The existing flow estimation and capillary segmentation in CycleTrack compute these values but additional work is required to quantitatively assess the accuracy of these estimations. Third, the proposed CenterNet based CycleTrack is also compatible with other object detectors. With the advances in object detections, more efficient and accurate detectors especially for small and crowded objects, like YOLO [32] and point-supervised deep detection network [33], can improve accuracy and speed to achieve real-time cell tracking on OBC videos. A new research has demonstrated that the non-invasive capillary videos can be acquired by a lower-cost reverse lens attachment to a mobile phone camera [34]. Therefore, a lightweight tracking algorithm may enable such portable capillaroscopy system more likely to become a practical point-of-care solution. Fourth, beyond constituent blood cell counts, OBC has the potential to measure Mean Corpuscular Volume (MCV), which measures average size of RBCs, Red cell Distribution Width (RDW), which quantifies RBC size variations, and other morphological biomarkers. To accomplish this, algorithm with an instance segmentation branch, like YOLACT [35] or CenterMask [36] could be employed. Though it would be advantageous for these analytics to also run quickly, it is also possible that the blood count is computed with efficient trackers and the analysis that requires accurate segmentation is computed as a subsequent step or via cloud computing.

## V. Conclusion

We presented a deep tracking model, called CycleTrack, that automatically counts blood cells from oblique back-illumination capillaroscopy videos. CycleTrack combines two state-of-the-art online tracking models, SORT and CenterTrack, and predicts back-and-forth cell displacement vectors to achieve optimal matching between newly detected cells and previously tracked cells in two consecutive frames with minimal increase in runtime. We make two simple assumptions about blood flow that enhance the accuracy of our model: (1) cells in the same capillary tend to flow in similar directions in a single frame, and (2) individual cell moves with a roughly linear constant velocity across frames. CycleTrack results outperform four existing multi-object tracking models on OBC videos and also demonstrates robust and accurate cell counting. In addition, CycleTrack is a promising model to explore other valuable clinical biomarkers from OBC videos, like blood velocity and heartrate.

### Acknowledgement

This research was supported in part with a gift from Fifth Generation, Inc.

Preprint 9